%% file: emnlp2020.tex
\documentclass[11pt,a4paper]{article}
\usepackage[hyperref]{acl2021}
\usepackage{times}
\usepackage{latexsym}
\usepackage{marvosym}
\usepackage{blindtext}
\usepackage{amsmath,amsthm,amssymb,pstricks,pst-node}
\usepackage{xspace,textcomp}
\usepackage{subcaption}
\usepackage{graphicx}
\usepackage{mathtools}
\usepackage{nicefrac}
\usepackage{bbm}

\usepackage{microtype}
\usepackage{footnote}
\usepackage{afterpage,refcount}

\newenvironment{enumeratesquish}{\begin{list}{\addtocounter{enumi}{1}\labelenumi}{\setlength{\itemsep}{0em}\setlength{\labelwidth}{0.5em}\setlength{\leftmargin}{\labelwidth}\addtolength{\leftmargin}{\labelsep}}}{\end{list}\setcounter{enumi}{0}}

\aclfinalcopy %
\def\aclpaperid{221} %

\newcommand{\kevin}[1]{\textcolor{blue}{\bf \small [ #1 --KG]}}
\newcommand{\mingda}[1]{\textcolor{red}{\bf \small [ #1 --MC]}}
\newcommand{\sam}[1]{\textcolor{magenta}{\bf \small [ #1 --SW]}}

\renewcommand{\kevin}[1]{}
\renewcommand{\mingda}[1]{}
\renewcommand{\sam}[1]{}

\newcommand{\dataset}{\textsc{WikiTableT}\xspace}
\newcommand{\wikidata}{Wikidata\xspace}
\newcommand{\wikibio}{\textsc{WikiBio}\xspace}
\newcommand{\entmax}{$\alpha$-entmax\xspace}

\newcommand{\bs}[1]{\boldsymbol{#1}}

\title{wiki2data2text}
\title{Large-Scale Generation of Wikipedia Article Sections from Diverse Data Sources}

\title{Generating Wikipedia Article Sections from Diverse Data Sources}
\title{\dataset: A Large-Scale Data-to-Text Dataset\\for Generating Wikipedia Article Sections}
\author{Mingda Chen\qquad  Sam Wiseman\qquad Kevin Gimpel\\
Toyota Technological Institute at Chicago, Chicago, IL, 60637, USA\\
  {\tt \{mchen,swiseman,kgimpel\}@ttic.edu}\\}

\date{}

\begin{document}
\maketitle
\begin{abstract}
Datasets for data-to-text generation typically focus either on multi-domain, single-sentence generation or on single-domain, long-form generation. In this work, we cast generating Wikipedia sections as a data-to-text generation task and create a large-scale dataset, \dataset, that pairs Wikipedia sections with their corresponding tabular data and various metadata. \dataset contains millions of instances, covering a broad range of topics, as well as a variety of flavors of generation tasks with different levels of flexibility. We benchmark several training and decoding strategies on \dataset.
Our qualitative analysis shows that the best approaches can generate fluent and high quality texts but they %
struggle with coherence and factuality, showing the potential for our dataset to inspire future work on long-form generation.\footnote{Code, data, and pretrained models are available at \url{https://github.com/mingdachen/WikiTableT}}
\end{abstract}

\section{Introduction}
\input{intro}
\section{Related Work}
\input{related_work}

\section{The \dataset Dataset}
\input{dataset}
\section{Methods}
\input{method}

\section{Experiments}
\input{experiment}

\section{Analysis}
\input{analysis}

\section{Conclusion}
\input{conclusion}
\section*{Acknowledgments}
This work was supported in part by a Google Fellowship to M. Chen. %

\section*{Impact Statement}
\input{impact}

\bibliographystyle{acl_natbib}
\bibliography{anthology,emnlp2020}
\newpage
\input{appendix}

\end{document}

%% file: intro.tex
Data-to-text generation \cite{kukich-1983-design,mckeown1992text} is the task of generating text based on structured data. Most existing data-to-text datasets focus on single-sentence generation, such as \wikibio \cite{lebret-etal-2016-neural}, LogicNLG \cite{chen-etal-2020-logical}, and ToTTo \cite{parikh-etal-2020-totto}. Other datasets are relatively small-scale and focus on long-form text generation, such as \textsc{RotoWire} \cite{wiseman2017challenges} and MLB \cite{puduppully-etal-2019-data}. In this work, we cast generating Wikipedia sections as a data-to-text generation task and build a large-scale dataset targeting multi-sentence data-to-text generation with a variety of domains and data sources.

To this end, we create a dataset that we call \dataset (``Wikipedia Tables to Text'') that pairs Wikipedia sections with their corresponding tabular data and various metadata. The data resources we consider are relevant either to entire Wikipedia articles, such as Wikipedia infoboxes and Wikidata tables, or to particular sections. Data from the latter category is built automatically from either naturally-occurring hyperlinks or from named entity recognizers. This data construction approach allows us to collect large quantities of instances while still ensuring the coverage of the information in the table. We also perform various types of filtering to ensure dataset quality. %

\dataset contains millions of instances  covering a broad range of topics and a variety of flavors of generation with different levels of flexibility. Figure~\ref{fig:dataset-example} shows two examples from \dataset. The first instance has more flexibility as it involves generating a fictional character biography in a comic book, whereas the second is more similar to standard data-to-text generation tasks, where the input tables contain all of the necessary information for generating the text. While the open-ended instances in \dataset are to some extent similar to story generation \cite{propp1968morphology,mcintyre-lapata-2009-learning,fan-etal-2018-hierarchical}, the fact that these instances are still constrained by the input tables enables different evaluation approaches and brings new challenges (i.e., being coherent and faithful to the input tables at the same time). %

Because of the range of knowledge-backed generation instances in \dataset, models trained on our dataset can be used in assistive writing technologies for a broad range of topics and types of knowledge.
For example, technologies can aid students in essay writing by drawing from multiple kinds of factual sources.
Moreover, \dataset can be used as a pretraining dataset for other relatively small-scale data-to-text datasets (e.g., \textsc{RotoWire}). A similar idea that uses data-to-text generation to create corpora for pretraining language models has shown promising results \citep{agarwal-etal-2021-knowledge}.

In experiments, we train several baseline models on \dataset and empirically compare training and decoding strategies.
We find that the best training strategies still rely on enforcing hard constraints to avoid overly repetitive texts. Human evaluations reveal that (1) humans are unable to differentiate the human written texts from the generations from our neural models; (2) while the annotations show that grammatical errors in the reference texts and the generations %
may prevent humans from fully understanding the texts, the best decoding strategy (i.e., beam search with $n$-gram blocking \citep{paulus2018a})
does not have such a problem and shows the best performance on several aspects; (3) the degree of topical similarity between the generations and the reference texts depends on the open-endedness of the instances.

Our analysis %
shows that the generations are fluent and generally have high quality, but the models sometimes struggle to generate coherent texts for all the involved entities, suggesting future research directions. For example, when the instance has a high degree of flexibility, we find the models making mistakes about what a particular entity type is capable of. We also find errors in terms of the factuality of the generated text, both in terms of contradictions relative to the tables and commonsense violations.

%% file: related_work.tex
There have been efforts in creating data-to-text datasets from various resources, including sports summaries \cite{wiseman2017challenges,puduppully-etal-2019-data}, weather forecasts \cite{liang-etal-2009-learning}, and commentaries \cite{david2008sport}. Most of the recent datasets focus on generating single sentences given tables, such as \wikibio, ToTTo, LogicNLG, and WikiTableText \cite{bao2018table}, or other types of data formats, such as data triples \cite{Pavlos2017neural,gardent-etal-2017-webnlg,nan-etal-2021-dart}, abstract meaning representations \cite{flanigan-etal-2016-generation}, minimal recursion semantics \cite{hajdik-etal-2019-neural}, or a set of concepts \cite{lin-etal-2020-commongen}.
Other than single sentences, there have been efforts in generating groups of sentences describing humans and animals \cite{wang-etal-2018-describing}, and generating a post-modifier phrase for a target sentence given a sentence context \cite{kang-etal-2019-pomo}. 
In this work, our focus is long-form text generation and we are interested in automatically creating a large-scale dataset containing multiple types of data-to-text instances. 
As shown in Table \ref{tab:data-stats}, \dataset differs from these datasets in that it is larger in scale and contains multi-sentence texts. More details are in the next section.

Wikipedia has also been used to construct datasets for other text generation tasks, such as generating Wikipedia movie plots \cite{orbach-goldberg-2020-facts2story,rashkin-etal-2020-plotmachines} and short Wikipedia event summaries \cite{gholipour-ghalandari-etal-2020-large}, and summarizing Wikipedia documents  \cite{zopf-2018-auto,j2018generating} or summaries of aspects of interests \cite{Hayashi2020WikiAspAD} from relevant documents.

As part of this work involves finding aligned tables and text, it is related to prior work on aligning Wikipedia texts to knowledge bases \cite{elsahar-etal-2018-rex,logan-etal-2019-baracks}.

%% file: dataset.tex
\begin{figure*}[t]
    \centering
    \begin{subfigure}[t]{1.0\textwidth}
    \centering
    \includegraphics[scale=0.46]{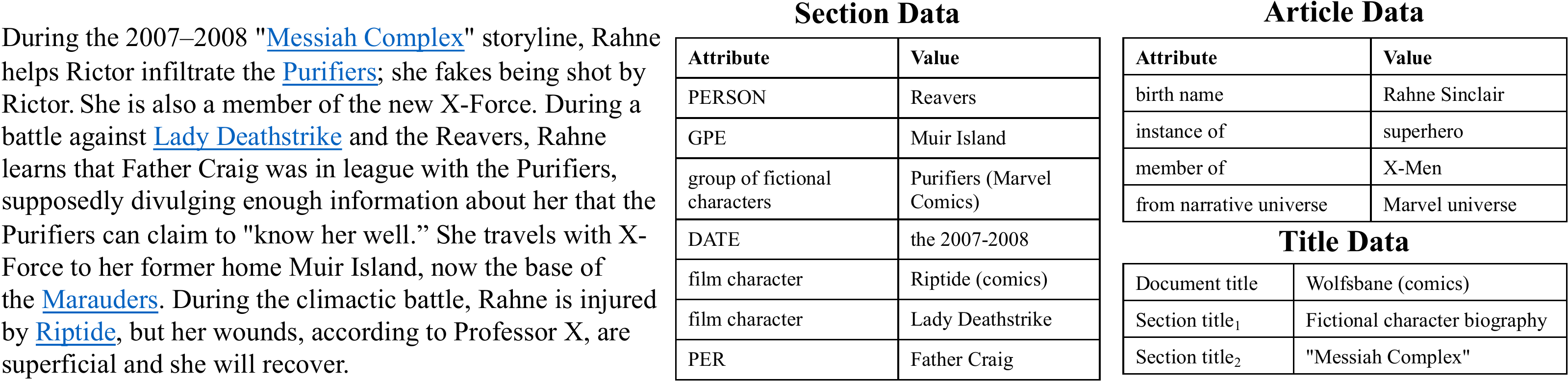}
    \end{subfigure}
    \begin{subfigure}[t]{1.0\textwidth}
    \vspace{0.1em}
    \centering
    \includegraphics[scale=0.46]{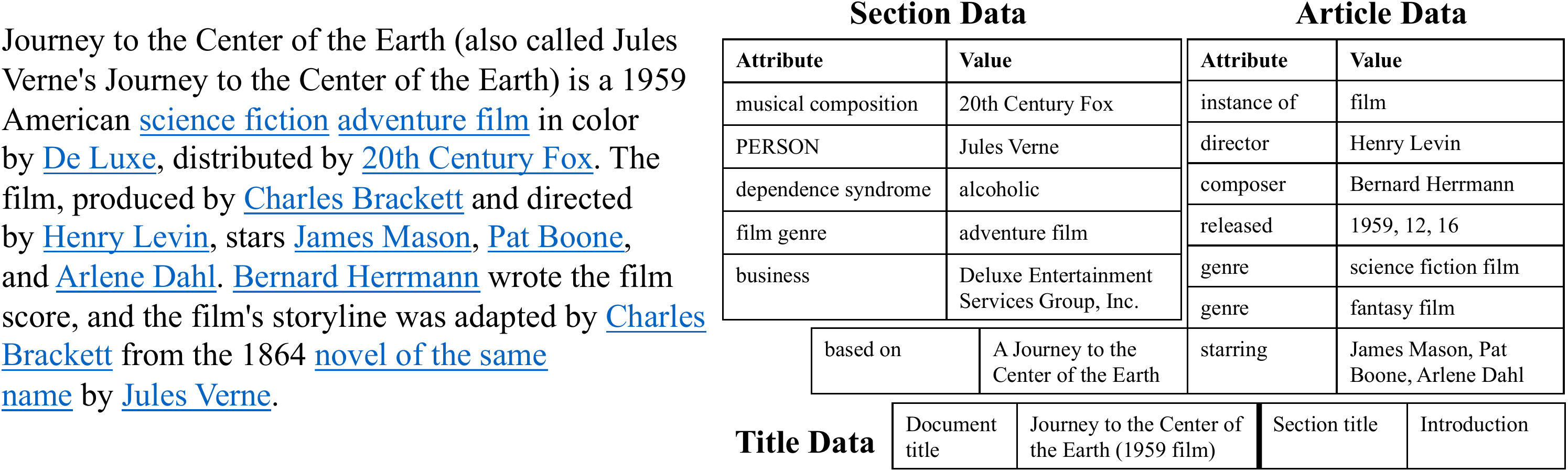}
    \end{subfigure}
    \caption{Two examples from \dataset. Only parts of the tables are shown due to space constraints. Underlined texts are hyperlinks. Records with the attributes ``DATE'', ``PER'', ``PERSON'',
    or ``GPE'' are from NER. The subscripts for section titles indicate the ordering of nesting, where smaller numbers are for higher level sections.
    }
    \label{fig:dataset-example}
\end{figure*}

The \dataset dataset pairs Wikipedia sections\footnote{We define a Wikipedia section to be all text starting after a (sub)section heading and proceeding until the next (sub)section heading. %
We include Wikipedia sections at various nesting levels. For example, a top level section may start with a few paragraphs describing general information followed by two subsections with more specific information, in which case the example will be converted into three instances in our dataset.} with their corresponding tabular data and various metadata; some of this data is relevant to entire Wikipedia articles (``article data'') or article structure (``title data''), while some is section-specific (``section data''). Each data table consists of a set of \textbf{records}, each of which is a tuple containing an \textbf{attribute} and a \textbf{value}.

The instances in \dataset cover a range of flavors of language generation. Some have more flexibility, requiring models to generate coherent stories based on the entities and knowledge given in the tables. The first instance in Figure~\ref{fig:dataset-example} is such an example. The text is from the Wikipedia article entitled ``Wolfsbane (comics)'' and resides within two nested sections: the higher-level section ``Fictional character biography'' and the lower-level section ``Messiah Complex''. The task is challenging as models need to generate a coherent passage that can connect all the entities in the section data, and the story also needs to fit the background knowledge provided in the article data.

Other instances are more similar to standard data-to-text generation tasks, where the input tables contain all the necessary information for generating the text. The second instance in Figure~\ref{fig:dataset-example} is an example of this sort of task. However, these tasks are still challenging due to the wide variety of topics contained in \dataset. %

\subsection{Dataset Construction}

We begin by describing the steps we take to construct \dataset. More details are in the supplementary material. In general, the steps can be split into two parts: collecting data tables and filtering out texts.
When collecting data, we consider five resources: \wikidata tables, infoboxes in Wikipedia pages,\footnote{Wikidata is a consistently-structured knowledge base (e.g., has a fixed set of attributes), whereas infoboxes are not consistently-structured and this flexibility sometimes allows the infobox to contain extra information. Therefore, we consider using infoboxes as extra resources.} hyperlinks in the passage, named entities in the passage obtained from named entity recognition (NER), and Wikipedia article structure.
For a given Wikipedia article, we use the same infobox and \wikidata table for all sections. These tables can serve as background knowledge for the article.
For each section in the article, we create a second table corresponding to section-specific data, i.e., section data.
The section data contains records constructed from hyperlinks and entities identified by a named entity recognizer.\footnote{We use the NER tagger from spaCy \cite{spacy2} and a BERT model  \cite{devlin-etal-2019-bert} finetuned on CoNLL03 data \cite{tjong-kim-sang-de-meulder-2003-introduction}.}

We form records for named entities by using the type of the entity as the attribute and the identified entity as the value. We form records for hyperlinks as follows. For the attribute, for a hyperlink with surface text $t$ and hyperlinked article $\ell$, we use the value of the ``instance of'' or ``subclass of'' tuple in the \wikidata table for $\ell$. For example, the first instance in Figure \ref{fig:dataset-example} will be turned into a record with attribute ``superhero'' and value ``Wolfsbane (comics)''.
If $\ell$ does not have a \wikidata table or no appropriate tuple, we consider the parent categories of $\ell$. For the value of the tuple, we use the document title of $\ell$ rather than the actual surface text $t$ to avoid giving away too much information in the reference text.

Complementary to the article data, we create a title table that provides information about the position in which the section is situated, which includes the article title and the section titles for the target section. As the initial sections in Wikipedia articles do not have section titles, we use the section title ``Introduction'' for these.

We also perform various filtering to ensure the quality of the data records, the coverage of the input data, and the length of the reference text. The final dataset contains approximately 1.5 million instances. We randomly sample 4533 instances as the development set and 4351 as the test set. We also ensure that there are no overlapping Wikipedia articles among splits.

\begin{table*}[t]
    \centering
    \small
\begin{tabular}{|l|ccccccc|}
\hline
& Vocab. & Tokens & Examples & Avg. Len. & Record Types & Avg. Records & Domain \\
\hline
WikiTableText & - & 185.0k & 13.3k & 13.9 & 3.0k & 4.1 & Wikipedia \\
\textsc{WikiBio} & 400.0k & 19.0M & 728.0k & 26.1 & 1.7k & 19.7 & Biography \\
\textsc{RotoWire} & 11.3k & 1.6M & 4.9k & 337.1 & 39.0 & 628.0 & Sports \\
MLB & 38.9k & 14.3M & 26.3k & 542.1 & 53.0 & 565.0 & Sports \\
LogicNLG & 122.0k & 52.7k & 37.0k & 14.2 & 11.7k & 13.5 & Wikipedia \\
ToTTo & 136.0k & 1.3M & 136.0k & 17.4 & 41.8k & 32.7 & Wikipedia \\
DART & 33.2k & 717.1k  & 82.2k & 21.6 & - & - & Wikipedia+Restaurant\\
\dataset & 1.9M & 169.0M & 1.5M$^\ast$
& 115.9 & 147.4k$^\dag$
& 51.9 & Wikipedia \\
\hline
\end{tabular}
\caption{
Statistics for several data-to-text datasets. \dataset combines a large number of examples, moderate generation length (typically more than one sentence), and a large variety of record types.
We omit record types and avg.~records for DART as its input units are triple sets instead of table records. $^\ast$887.7k unique Wikipedia articles. $^\dag$Number of record types for each resource: 31.8k (Infobox), 1.7k (\wikidata), 115.6k (Hyperlinks), 17 (NER).
}
    \label{tab:data-stats}
\end{table*}

\subsection{Dataset Characteristics}
Table~\ref{tab:data-stats} shows statistics for \dataset and related datasets. While the average length of a \dataset instance is not longer than some of the existing datasets, \dataset offers more diverse topics than the sports-related datasets \textsc{RotoWire} and MLB, or the biography-related dataset \wikibio. Compared to the prior work that also uses Wikipedia for constructing datasets, \wikibio, LogicNLG, ToTTo, and DART \cite{nan-etal-2021-dart} all focus on sentence generation, whereas \dataset requires generating Wikipedia article sections, which are typically multiple sentences and therefore more challenging. \dataset is also much larger than all existing datasets.

\begin{table}[t]
    \centering
    \small
\begin{tabular}{|l|c|}\hline
Category & Fraction (\%) \\\hline
human & 45.62 \\
film & \phantom{1}4.61 \\
single (music) & \phantom{1}1.74 \\
human settlement & \phantom{1}1.53 \\
album & \phantom{1}1.41 \\
sports season & \phantom{1}1.26 \\
television series & \phantom{1}1.17 \\
village & \phantom{1}1.12 \\
taxon & \phantom{1}0.89 \\
\hline
\end{tabular}

    \caption{Top 10 most frequent article categories and their corresponding proportions in \dataset.}
    \vspace{-1em}
    \label{tab:data-doc-cat-freq}
\end{table}

To %
demonstrate the diversity of topics covered in \dataset, we use either the ``instance of'' or ``subclass of'' relation from \wikidata as the category of the article.\footnote{When there are multiple values in these two relations, we pick the one that has the smallest number of words, as it often is the most generic phrase, suitable for representing the topic.}
We show the top 10 most frequent document categories in Table~\ref{tab:data-doc-cat-freq}. Due to the criteria we use for filtering, only 1.05\% of articles in \dataset do not have these relations or \wikidata entries, and we omit these articles in the table. As the table demonstrates, more than 50\% of the articles in \dataset are not about people (i.e., the topic of \wikibio), within which the most frequent category covers only 4.61\%.

\subsection{Dataset Challenges}
In this subsection, we highlight two challenges of \dataset.

\begin{enumeratesquish}
    \item In contrast to work on evaluating commonsense knowledge in generation where reference texts are single sentences describing everyday scenes \cite{lin-etal-2020-commongen}, \dataset can serve as a testbed for evaluating models’ abilities to use world knowledge for generating coherent long-form text.
    \item Compared to other long-form data-to-text datasets such as \textsc{RotoWire} where the input tables are box scores, the input tables in \dataset are more diverse, including both numbers (e.g., economy and population data of an area throughout years), and short phrases. This makes \dataset more challenging and applicable to various scenarios.

\end{enumeratesquish}

%% file: method.tex
In this section, we describe details of models that we will benchmark on \dataset.

Our base model is based on the transformer \cite{attn_is_all_you_need}. To encode tables, we linearize the tables by using special tokens to separate cells and using feature embeddings to represent records in tables. 
For the title table in the first instance in Figure~\ref{fig:dataset-example} 
the linearized table will be
\begin{equation}
\begin{aligned}
    &\langle\text{boc}\rangle_1 \text{Doc.}_1\text{ title}_1 \langle\text{bov}\rangle_1 \text{ Wolfsbane}_1\text{ (comics)}_1\\ &\langle\text{boc}\rangle_2 \text{Sec.}_2\text{ title}_2 \langle\text{bov}\rangle_2 \text{ Fictional}_2\text{ character}_2\\ &\text{ biography}_2\langle\text{boc}\rangle_3 \cdots \langle\text{eoc}\rangle
\end{aligned}
\label{eq:linear-table}
\end{equation}
As shown in Eq.~\ref{eq:linear-table}, we employ several techniques when encoding tables: (1) we use special tokens $\langle\text{boc}\rangle$ and $\langle\text{bov}\rangle$ to separate attributes and values, and  $\langle\text{eoc}\rangle$ to indicate the end of a sequence; (2) we use subscript indices to indicate unique ID embeddings that are added to the embeddings for each record, which helps models align attributes with values; and (3) we restart the positional embeddings at each $\langle\text{boc}\rangle$, such that models will not use the ordering of the input records. In addition, we add a special embedding to each record to indicate if it is from the section table or the article/title table. In \wikidata, there could be multiple qualifiers attached to a record, in which case we replicate the record for each qualifier separately.

Similar linearization approaches have been used in prior work \cite{dhingra-etal-2019-handling,hwang2019comprehensive,herzig-etal-2020-tapas,yin-etal-2020-tabert}.
With linearized tables, training and inference become similar to other sequence-to-sequence settings.
We train our models with teacher-forcing and standard cross entropy loss unless otherwise specified.

\subsection{Training Strategies}

We experiment with three types of modifications to standard sequence-to-sequence training: %

\paragraph{\entmax.} \entmax~\cite{peters-etal-2019-sparse} is a mapping from scores to a distribution that permits varying the level of sparsity in the distribution. This mapping function has been used in machine translation~\cite{peters-etal-2019-sparse} and text generation~\cite{martins-etal-2020-sparse}. When using \entmax in the decoder, we also replace the cross entropy loss with the \entmax loss~\cite{peters-etal-2019-sparse}. Both \entmax and the \entmax loss have a hyperparameter $\alpha$. We follow \citet{martins-etal-2020-sparse} and use $\alpha=1.2$ as they found it to be the best value for reducing repetition in generation.

\paragraph{Copy Mechanism.} Similar to prior work on data-to-text generation \cite{wiseman2017challenges,puduppully-etal-2019-data}, we use pointer-generator network style copy attention \cite{see-etal-2017-get} in the decoder.

\paragraph{Cyclic Loss.}
Cyclic losses have been shown to be effective in textual style transfer \cite{shetty2018a4nt,pang-gimpel-2019-unsupervised} 
and neural machine translation~\cite{cheng-etal-2016-semi,dual_nips2016,tu2017neural}. 
\citet{wiseman2017challenges} also used this for data-to-text and found it helpful for generating long sequences. In this work, we experiment with adding the cyclic loss to our transformer models, where the backward model can be seen as an information extraction system. We expect that adding the cyclic loss should enable a data-to-text model to generate sentences that are more faithful to the conditioned tables. 
The cyclic loss is used during training only and does not affect the models during inference. More details are in the appendix. %

\subsection{Decoding Strategies}

\citet{massarelli2019decoding} showed that the choice of decoding strategy can affect the faithfulness or repetitiveness of text generated by language models. %
We are also interested in these effects in the context of data-to-text generation, and therefore benchmark several decoding strategies on \dataset. 
Our models use byte-pair encoding (BPE; \citealp{sennrich-etal-2016-neural}) and for all of the following strategies, we always set the minimum number of decoding steps to 100 as it improves most of the evaluation metrics, and the maximum number of decoding steps to 300.

Specifically, we benchmark (1) greedy decoding; (2) nucleus sampling \cite{Holtzman2020The} with threshold 0.9 as suggested by \citet{Holtzman2020The}; (3) beam search; and (4) beam search with $n$-gram blocking \citep{paulus2018a} where we set the probabilities of repeated trigrams to be 0 during beam search. We set the beam size %
to be 5 by default.
The appendix has more details about the decoding strategies.

%% file: experiment.tex
\subsection{Setup}

\begin{table*}[t]
    \centering
    \small
    \begin{tabular}{|l|c|c|c|c|c|c|c|}\hline
& REP & BLEU & RL & MET & PAR-P & PAR-R & PAR-F1 \\\hline
References & 1.2 & 100.0 & 100.0 & 100.0 & 100.0 & 59.2 & 72.9 \\
Linearized article tables & 8.0 & 2.2 & 14.7 & 9.3 & 100.0 & 16.3 & 25.6 \\
Linearized section tables & 1.0 & 1.9 & 27.9 & 15.5 & 100.0 & 20.9 & 33.4 \\
Linearized tables & 7.9 & 6.4 & 22.0 & 18.3 & 100.0 & 48.3 & 63.0 \\
Linearized tables + references & 7.6 & 36.5 & 61.3 & 56.5 & 99.9 & 100.0 & 100.0 \\\hline
\multicolumn{8}{c}{ Base models trained on the 500k training set (beam search) } \\\hline
Base & 33.0 & 15.6 & 36.9 & 20.3 & 66.3 & 28.8 & 37.7 \\
Base + entmax & 25.9 & 15.4 & 36.2 & 20.3 & 64.6 & 29.0 & 37.7 \\
Base + copy & 30.1 & 15.9 & 37.5 & 20.7 & 67.1 & 29.4 & 38.5 \\
Base + copy + cyclic loss & 28.0 & 15.7 & 37.5 & 20.8 & 67.5 & 29.7 & 38.9 \\\hline
\multicolumn{8}{c}{ Large models trained on the full training set (different decoding strategies) } \\\hline
Large + greedy & 26.8 & 18.9 & 38.5 & 23.5 & 60.4 & 33.1 & 40.4 \\
Large + nucleus sampling & 2.3 & 18.3 & 36.1 & 23.7 & 54.2 & 32.5 & 38.7 \\
Large + beam search & 18.8 & 19.5 & 39.9 & 23.9 & 65.8 & 34.3 & 42.8 \\
Large + beam search + $n$-gram blocking & 1.9 & 19.3 & 39.3 & 24.4 & 62.2 & 35.3 & 43.0 \\\hline
\end{tabular}

    \caption{Test set results for our models. When training the large models, we use the ``copy + cyclic loss'' setting as it gives the best performance for the base models for most of the metrics.}
    \vspace{-1em}
    \label{tab:main-result}
\end{table*}

We experiment with two sizes of transformer models. One is ``Base'', where we use a 1-layer encoder and a 6-layer decoder, each of which has 512 hidden size and 4 attention heads. The other one is ``Large'', where we use a 1-layer encoder and a 12-layer decoder, each of which has 1024 hidden size and 8 attention heads. Models similar to the base configuration have shown strong performance on \textsc{RotoWire}~\cite{gong-etal-2019-enhanced}.\footnote{When training the base model with entmax on \wikibio, it achieves BLEU-4 45.75 and ROUGE-4 39.39 on the test set using greedy decoding, which are comparable to the current state-of-the-art results of~\citet{liu2017aaai}.} Due to limited computational power, we parameterize our backward model as a transformer model with a 2-layer encoder and a 2-layer decoder.\footnote{We did not experiment with pretrained models because they typically use the entirety of Wikipedia, %
which would presumably overlap with our test set.}

We use BPE with 30k merging operations. We randomly sample 500k instances from the training set and train base models on them when exploring different training strategies. We train a large model with the best setting (using the copy mechanism and cyclic loss)  on the full training set. We train both models for 5 epochs. During training we perform early stopping on the development set using greedy decoding.

We report BLEU \cite{papineni-etal-2002-bleu}, ROUGE-L (RL) \cite{lin-2004-rouge}, METEOR (MET) \cite{banerjee-lavie-2005-meteor}, and PARENT \cite{dhingra-etal-2019-handling}, including precision (PAR-P), recall (PAR-R), and F1 (PAR-F1) scores. The first three metrics consider the similarities between generated texts and references, whereas PARENT also considers the similarity between the generation and the table. %
When using PARENT, we use all three tables, i.e., the section, article, and title tables.

As we are also interested in the repetitiveness of generated texts, we define a metric based on $n$-gram repetitions which we call ``REP''. REP computes the ratio of the number of repeated $n$-grams to the total number of $n$-grams within a text, so when REP has higher value, it indicates that the text has more repetitions. Here we consider $n$-grams that appear 3 or more times as repetitions and the $n$-grams we consider are from bigrams to 4-grams. When reporting REP scores for a dataset, we average the REP scores for each instance in the dataset. 
Similar metrics have been used in prior work \cite{Holtzman2020The,Welleck2020Neural}.

\subsection{Results}

In Table \ref{tab:main-result}, we report the test results for both our base models and large models. We also report a set of baselines that are based on simply returning the linearized tables and their concatenations with the references. The linearized table baselines show how much information is already contained in the table, while the reference baselines show the upper bound performance for each metric.

In comparing training strategies, we find that using \entmax improves REP significantly but not other metrics.
Adding the cyclic loss or the copy mechanism helps improve performance for the PAR scores and REP, %
and combining both further improves these metrics.

When comparing decoding strategies, we find that both nucleus sampling and $n$-gram blocking are effective in reducing repetition. 
Nucleus sampling harms the PAR scores, especially PAR-P, but has less impact on the other metrics, 
indicating that it makes the model more likely to generate texts that are less relevant to the tables. 
Using beam search improves all metrics significantly when compared to greedy decoding, especially the PAR-P and REP scores. Adding $n$-gram blocking further reduces the REP score, pushing it to be even lower than that from nucleus sampling, but still retains the improvements in PAR scores from beam search. The best overall decoding strategy appears to be beam search with $n$-gram blocking.

%% file: analysis.tex
We now describe a manual evaluation and analyze some generated examples. All results in this section use the development set. We also conduct experiments on analyzing the effect of using the section data and the article data during training, finding that the benefits that they bring to the model performance are complementary. See the appendix for more details.

\subsection{Human Evaluation}

We conduct a human evaluation using generations from the large model on the development set. We choose texts shorter than 100 tokens and that cover particular topics
as we found during pilot studies that annotators struggled with texts that were very long or about unfamiliar topics.\footnote{We did not find the filtering to change the observed trends for the automatic metrics and provide the list of selected topics in the appendix.} 

We design two sets of questions. The first focuses on the text itself (i.e., grammaticality and coherence) and its faithfulness to the input article table. Since this set does not involve the reference, we can ask these questions about both generated texts and the reference texts themselves. 
The second set of questions evaluates the differences between the generations and the reference texts (i.e., relevance and support), allowing us to see if the generated text matches the human written section text. Specifically, relevance evaluates topical similarity between generations and references, and support evaluates whether the facts expressed in the generations are supported by or contradictory to those in the references. %
The full questions and numerical answer descriptions %
are in the appendix.  %

\begin{table}[t]
    \centering
    \small\setlength{\tabcolsep}{4pt}
\begin{tabular}{|l|c|c|c|}\hline
 & Grammar & Coherence & Faithfulness \\\hline
Reference & 4.0 (1.0) & 4.1 (0.9) & 3.8 (0.8) \\\hline
Beam search & 4.0 (1.0) & 4.0 (1.0) & 3.9 (1.0) \\
Nucleus sampling & 4.0 (0.8) & 4.1 (0.9) & 3.9 (0.8) \\
$n$-gram blocking & 4.2 (0.9) & 4.2 (0.9) & 3.9 (1.0) \\
\hline
\end{tabular}
    \caption{Average human ratings (standard deviations in parentheses) for %
    grammaticality, coherence, and faithfulness to the input article table. }
    \label{tab:human_eval_table}
\end{table}

\begin{table}[t]
    \centering
    \small\setlength{\tabcolsep}{5pt}
\begin{tabular}{|l|c|c|}\hline
 & Relevance & Support \\\hline
Beam search & 3.8 (1.1) & 3.6 (1.2) \\
Nucleus sampling & 3.7 (1.2) & 3.8 (1.1) \\
$n$-gram blocking & 3.9 (1.0) & 3.8 (1.0) \\
\hline
\end{tabular}
    \caption{Average human ratings (standard deviations in parentheses) 
    of relevance and support when comparing to the reference text.
    }
    \vspace{-1.5em}
    \label{tab:human_eval_reference}
\end{table}

We report results in Tables \ref{tab:human_eval_table} and  \ref{tab:human_eval_reference}. The scores are on a 1-5 scale with 5 being the best. For the first set, we collect 480 annotations from 38 annotators. For the second set, we collect 360 annotations from 28 annotators. We also ensure that each system has the same number of annotations.\footnote{We used Amazon Mechanical Turk. To ensure annotation quality, we only recruited annotators with master qualification. We collected one annotation for each instance (so that we can cover more instances) and paid 30 cents per annotation. The amount of wage per annotation is decided by (1) the amount of time each annotator spent on the task during our pilot study and (2) a target hourly wage of approximately \$11.}

\begin{table*}[t]
    \centering\small\setlength{\tabcolsep}{4pt}
\begin{tabular}{|p{0.08\textwidth}|p{0.815\textwidth}|c|c|}\hline
\multicolumn{1}{|c|}{Method} &\multicolumn{1}{c|}{Text} & G & C \\\hline

Reference & He contested the parliamentary seat of Meriden at the 1987 general election, where he was defeated by the sitting Conservative MP Iain Mills by a margin of 16,820. He was then selected to fight the Conservative-held marginal seat of Birmingham Northfield ... & 3 & 3 \\\hline

Reference & Boscawen married on 23 April 1700 in Henry VII's Chapel, Westminster Abbey, Charlotte Godfrey elder daughter and coheir of Colonel Charles Godfrey, master of the jewel office and his wife Arabella Churchill ... & 3 & 4 \\\hline

Sampling & 7th Marquess of Exeter married, firstly, Edith Csanady de Telegd (born \textbf{1 September 1935} in England; died 16 June 1956 in London), on \textbf{17 January 1934} ... & 4 & 5 \\\hline

Blocking & ... He averaged 10.9 rebounds and 3.0 assists per game \textbf{as a senior in 1987-88}. He was selected to the Sweet 16 of the NCAA Tournament \textbf{as a junior in 1988-89} ... & 5 & 5 \\\hline
\end{tabular}

    \caption{Human annotation examples for grammaticality (G) and coherence (C). Due to space constraints, only parts of the texts are shown. We highlight texts that are incoherent.}
    \vspace{-1em}
    \label{tab:human_eval_examples}
\end{table*}

It is interesting to note from Table \ref{tab:human_eval_table} that human annotators are unable to differentiate the human written texts from the generations from our neural models. Since the Wikipedia section texts are parts of Wikipedia articles, showing the section texts in isolation can make them difficult to understand, potentially resulting in noisy annotations. As shown by the first instance in Table \ref{tab:human_eval_examples}, the text uses the pronoun ``he'' without clarifying what the pronoun refers to. The paragraph is rated 3 for coherence, presumably due to this ambiguity. Also, Wikipedia texts are sometimes grammatically complex and annotators can mistake them for being ungrammatical, e.g., the second instance in Table \ref{tab:human_eval_examples}. 

On the other hand, the coherence errors in the generated texts are not always easy to spot. See, for example, the last two instances in Table \ref{tab:human_eval_examples}, where the incoherence lies in the facts that (1) it is impossible to marry a person before the person is born, and (2) senior year takes place after junior year. These details are embedded in long contexts, which may be overlooked by annotators and lead to results favorable to these neural models.

To study the relationship between coherence and grammaticality, we compute Spearman's correlations between the human annotations for coherence and grammaticality after removing the ones with perfect scores for coherence. Table \ref{tab:human_eval_spearman} shows the results. %
The correlations are much higher for references, beam search, and nucleus sampling than for $n$-gram blocking. This trend suggests that the imperfect coherence scores for the reference texts are likely because annotators find the texts to contain grammatical errors (or to possess grammatical complexity) which may prevent them from fully understanding the texts. %
However, $n$-gram blocking does not have this problem and thus achieves the best results for both coherence and grammaticality. We hypothesize that $n$-gram blocking is able to avoid the types of grammatical errors that prevent understanding because (1) unlike nucleus sampling, $n$-gram blocking does not rely on randomness to avoid repetition; (2) $n$-gram blocking does not suffer from repetitions like beam search.

\begin{table}[t]
    \centering
    \small\setlength{\tabcolsep}{5pt}
\begin{tabular}{|l|c|c|c|c|}\hline
 & Ref. & Beam & Samp. & Block. \\\hline
Spearman corr. & 39.6 & 39.7 & 40.8 & 16.4 \\
\# annotations & 67 & 80 & 76 & 67  \\
\hline
\end{tabular}
    \caption{Spearman correlations between the human evaluation results for grammaticality and coherence. We omit annotations with perfect scores for coherence.}
    \label{tab:human_eval_spearman}
\end{table}

\begin{table}[t]
    \centering
    \small\setlength{\tabcolsep}{5pt}
\begin{tabular}{|l|c|c|c|c|c|}\hline
 & 1 & 2 & 3 & 4 & 5 \\\hline
Relevance & 24.2 & 19.2 & 13.6  & 12.0 & 8.9 \\
\# annotations & 10 & 48 & 65 & 124 & 113 \\\hline
Support & 17.0 & 11.0 & 17.5 & 12.5  & 9.4  \\
\# annotations & 13 & 47 & 68 & 135 & 97 \\
\hline
\end{tabular}
    \caption{Averaged perplexities and the corresponding numbers of annotations for each option for the relevance and support questions (5 is the best option). We aggregate annotations for different decoding algorithms. We note that the perplexities are computed based on the reference texts using the large model.}
    \vspace{-1em}
    \label{tab:human_eval_avg_ppl}
\end{table}

We report results for the second set of questions in Table \ref{tab:human_eval_reference}. The three evaluated systems show similar performance. To investigate the relationship between the degree of open-endedness of a \dataset instance and its corresponding evaluation scores, we compute the averaged perplexities (based on our large models) for each option in Table \ref{tab:human_eval_avg_ppl}. 
The most relevant generations are typically from more closed-ended or constrained instances.\footnote{\citet{li2015nlp} use entropy as a proxy to quantify complexity of tasks. In this work, we use perplexity to measure how open-ended the instances are.} Similarly for the support scores, more open-ended instances are distributed at score 3, which means that there is no fact supported by or contradictory to the shown tables. While the open-endedness of an instance usually depends on its topics (e.g., movie plots are open-ended), there are many cases where the models can benefit from better entity modeling, such as understanding what a particular entity type is capable of (e.g., see the last example in Sec. \ref{sec:generation_example}).

Recent work has also found conducting human evaluation for long-form generation to be challenging, for example in the context of 
question answering \cite{krishna-etal-2021-hurdles} and story generation \cite{akoury-etal-2020-storium}.  Our observations for data-to-text generation complement theirs and we hope that our dataset can inspire future research on human evaluation for long-form text generation.

\begin{table}[t]
    \centering\small
    \begin{tabular}{|l|r|r|}\hline
        percentile & train perp. & dev perp. \\\hline
        10 &  2.3 & 2.5 \\
        20 &  3.1 & 3.6 \\
        30 &  4.0 & 4.7 \\
        40 &  4.9 & 6.1 \\
        50 &  6.1 & 7.6 \\
        60 &  7.6 & 9.8 \\
        70 & 9.8 & 12.9 \\
        80 & 13.3 & 18.2 \\
        90 & 20.6 & 29.3 \\
        100 & 278.8 & 129.4 \\\hline
    \end{tabular}
    \caption{Percentiles of perplexities for training and development splits of our dataset. We use the large model.}
    \label{tab:dataset_ppl}
\end{table}

\begin{table*}[t]
    \centering\small\setlength{\tabcolsep}{5pt}
\begin{tabular}{|p{0.08\textwidth}|p{0.87\textwidth}|}\hline
\multicolumn{1}{|c|}{Method} &\multicolumn{1}{c|}{Generated Text} \\\hline

Nucleus sampling (threshold 0.9) & In a two-issue mini-series titled "Messiah Complex", portrayed by Maisie Williams, Father Craig is ordered to rescue the world from Reavers and run out of the X-Force causing the X-Force to permanently withdraw. The original humans were replaced by slots and Muir Island being pulled away. They also possessed the entire X-Force after issue \#9 in which the X-Force appeared. Claremont and Sinclair later reprised their parts in "The Donderful World of Wolverine" with Wolfsbane as one of the new X-Men and her group taken back to their reservation for unknown reasons. \\\hline
$n$-gram blocking & In the "Messiah Complex" storyline, Wolfsbane is a member of the X-Men, a group of mutants recruited by Father Craig to fight the Reavers in Muir Island. Father Craig is killed in the battle with Reavers, but Father Craig manages to rescue the mutant from the Rain of Gesham and retrieve the mutant. She is later revealed to have been killed by Rain in a battle with Father Craig. In the aftermath, Father Craig tells Father Craig that she is the only mutant to be killed by a mutant. \\\hline\hline

Nucleus sampling (threshold 0.9) & In May 2013, McCoy allegedly assault a Spoiler business official. The incident motivated him to take a business trip to Harrisburg, Pennsylvania and sign a letter from the Buffalo Bills paying him over \$50,000 as well as verbally prohibiting legal kickoff work. In September 2014, McCoy shared the issue with some former players of the first two film franchises. McCoy played a leading role in the 2012 Marvel Cinematic Universe film "Erab of the Press", and was named one of the film's biggest stars on Twitter. On September 9, 2014, McCoy filed a lawsuit against PYT to change its Twitter name to LeSean McCoy Productions. \\\hline
$n$-gram blocking & On September 9, 2014, McCoy was arrested in Harrisburg, Pennsylvania on suspicion of assault. He was charged with assault and battery. In May 2013, he was fined over \$50,000 by the Buffalo Bills. In September 2014, he was suspended for two games by the PYT for violating the Marvel Cinematic Universe. He was released by the Bills in October of the same year. He was cleared of all charges on Twitter, and was banned from playing in the 2014 Pro Bowl due to his Twitter account. \\\hline

\end{tabular}

    \caption{Generation examples from the large model. The first example corresponds to the first instance in Figure~\ref{fig:dataset-example}. The complete set of generations is in the appendix.}
    \vspace{-1em}
    \label{tab:generation_examples}
\end{table*}

\subsection{Distribution of Perplexity}

To determine the fraction of \dataset that can be seen as constrained, we report the percentiles of perplexities for training and development splits in Table~\ref{tab:dataset_ppl}. From Table~\ref{tab:human_eval_avg_ppl}, it can be observed that instances with perplexities around 9.0 generally lead to model generations that are closely relevant to the reference texts and mostly supported by the input tables, and therefore are likely to be the constrained instances. From Table~\ref{tab:dataset_ppl}, we see that at least half of our dataset has perplexities lower than 9.0, so we conjecture that half of our dataset consists of constrained instances.

\subsection{Generation Examples}
\label{sec:generation_example}

Table~\ref{tab:generation_examples} shows generation examples 
for nucleus sampling and beam search with $n$-gram blocking. We observe very different trends between the two instances in Figure~\ref{fig:dataset-example}. For the first instance about the X-Men, although both generations look fluent, their stories differ dramatically. The generated text for nucleus sampling describes a story that starts by saying Father Craig rescues the world from Reavers and ends with Wolfsbane joining as one of the new X-Men. On the other hand, $n$-gram blocking generates a story where Wolfsbane already is a member of X-Men, and the story says Father Craig fought and was killed by the Reavers, but manages to rescue the mutant. For the less open-ended instances (e.g., the second instance in Figure~\ref{fig:dataset-example}), different decoding strategies mostly generate similar details (see the appendix for generations).  %

Despite having different details, %
these generations appear to try to fit in as many %
entities from the tables as possible, in contrast to beam search (shown in the appendix) which mostly degenerates into repetition for more open-ended instances. This explains our previous observation that $n$-gram blocking helps with the PAR-R score.

Even though the generations are of good quality for most instances, their implausibility becomes more apparent when readers have enough background knowledge to understand the involved entities. For example, the second instance in Table \ref{tab:generation_examples} comes from the Wikipedia page ``LeSean McCoy'' (a football player) under the sections ``Personal life'' and ``Controversies'' (details in the appendix). 
The generation from nucleus sampling is implausible/nonsensical in some places (``assault a Spoiler business official'') and factually incorrect elsewhere (McCoy did not play a leading role in any film, and ``Erab of the Press'' is not an actual film). 
The fourth generation 
is implausible because a player is unlikely to be suspended for ``violating the Marvel Cinematic Universe'', and it is unlikely for a person to be cleared of all charges on Twitter. 
Our models have limited access to knowledge about entities, e.g., the capabilities of a social media company like Twitter. Future research may incorporate extra resources, make use of pretrained models, or incorporate factuality modules to solve these problems.

%% file: conclusion.tex
We created \dataset, a dataset that contains Wikipedia article sections and their corresponding  tabular  data  and  various metadata. \dataset contains millions of instances covering a broad range of topics and kinds 
of generation tasks. 
Our manual evaluation showed that humans are unable to differentiate the references and model generations, and $n$-gram blocking performs the best on grammaticality and coherence. However, qualitative analysis showed that our models sometimes struggle with coherence and factuality, suggesting several directions for future work. 

%% file: impact.tex
We highlight a few limitations as follows: (1) Wikipedia texts are generally written in objective tones, but some of the texts may contain controversial content that even the community contributors do not agree upon; (2) models trained on our dataset may generate deceitful texts that are unfaithful to what actually happened to particular entities; (3) though the instances in \dataset cover various topics, the writing style is almost always the same. Future work may explore more diverse writing styles.

%% file: appendix.tex
\appendix

\section{Dataset Construction}
When collecting data, we consider five resources: \wikidata tables, infoboxes in Wikipedia pages, hyperlinks in the passage, named entities in the passage obtained from named entity recognition (NER), and Wikipedia article structure.
For each article in Wikipedia, we use the same infobox and \wikidata table for all sections. These tables can serve as background knowledge for the article.
For each section in the article, we create a second table corresponding to section-specific data, i.e., section data.
The section data contains records constructed from hyperlinks and entities identified by a named entity recognizer. Section data contributes around 25\% of the records in \dataset.

We filter out several entity types related to numbers\footnote{List of filtered entity types: PERCENT, TIME, QUANTITY, ORDINAL, CARDINAL.} as the specific meanings of these numbers in the section of interest are difficult to recover from the information in the tables.
After filtering, we use the identified entities as the values and the entity types as the attributes. This contributes roughly 12\% of the records in our final dataset.

We also create records from hyperlinks in the section of interest. We first expand the hyperlinks available for each section with hyperlinks available in the parent categories. %
We first group hyperlinks across all Wikipedia articles with those same categories,
and then we perform string matching between these hyperlinks and the text in the section. If there are exact matches, we will include those hyperlinks as part of the hyperlinks in this section.

Details for constructing a record with attribute $a$ and value $v$ for a hyperlink with surface text $t$ and hyperlinked article $\ell$ are as follows.
To set $a$, we use the value of the ``instance of'' or ``subclass of'' tuple in the \wikidata table for $\ell$. If $\ell$ does not have a \wikidata table or no appropriate tuple, we consider the parent categories of $\ell$ as candidates for $a$. If there are multiple candidates for $a$, we first embed these candidates and $a$ using GloVe~\cite{pennington-etal-2014-glove} embeddings and then choose the one that maximizes cosine similarity between the document titles or section titles and the candidates for $a$.
For the value $v$ of the tuple, we use the document title of $\ell$ rather than the actual surface text $t$ to avoid giving away too much information in the reference text.  The records formed by hyperlinks contribute approximately 13\% of the records in \dataset.

We shuffle the ordering of the records from NER and the hyperlinks to prevent models from relying on the ordering of records in the reference text.

The records from the section data can be seen as section-specific information that can make the task more solvable. Complementary to the article data, we create a title table that provides information about the position in which the section is situated, which includes the article title and the section titles for the target section. As the initial sections in Wikipedia articles do not have section titles, we use the section title ``Introduction'' for these.\footnote{Among millions of section titles in Wikipedia, there are only 4672 sections, including nested sections, that are called ``Introduction''. Therefore, we believe this process will not introduce much noise into the dataset.}

As the records in our data tables come from different resources, we perform extra filtering to remove duplicates in the records. In particular, we give \wikidata the highest priority as it is a human-annotated well-structured data resource (infoboxes are human-annotated but not well-structured due to the way they are stored on Wikipedia) and the entities from NER the lowest priority as they are automatically constructed. That is, when we identify duplicates across different resources, we will keep the records from the higher priority resource and drop those from the lower one. More specifically, the duplicates between \wikidata records and infoboxes are determined by whether there are duplicate values or duplicate attributes: for hyperlinks and infoboxes or \wikidata, they are judged by duplicate values; for NER and hyperlinks, they are based on whether there is any token overlapping between values.

After table collection, we have the following criteria for filtering out the texts: (1) we limit the text length to be between 50 and 1000 word tokens; (2) to ensure that there is sufficient information in the table, we only keep data-text pairs that contain more than 2 records per sentence and more than 15 records per 100 tokens from \wikidata and infoboxes; (3) to avoid texts such as lists of hyperlinks, we filter out texts where more than 50\% of their word tokens are from hyperlink texts.

\section{Human Evaluation}

\begin{table}[t]
    \centering\small
    \begin{tabular}{|p{0.45\textwidth}|}\hline
        1 = it is completely ungrammatical, as it is impossible to understand the text. \\\hline
        2 = it has many grammatical errors, and these errors make the text very difficult to understand. \\\hline
        3 = it has grammatical errors, and some of them make part of the text difficult to understand. \\\hline
        4 = it has some grammatical errors, but they are minor errors that do not affect reading. \\\hline
        5 = it is completely grammatical, as it does not have any grammatical errors.\\\hline
    \end{tabular}
    \caption{Rating explanations %
    for grammaticality.}
    \label{sup:tab:human_evaluation_options_grammar}
\end{table}

\begin{table}[t]
    \centering\small
    \begin{tabular}{|p{0.45\textwidth}|}\hline
        1 = it is completely incoherent, as it is impossible to piece together information in the text. \\\hline
        2 = it is incoherent in most places. You can only understand part of the story. \\\hline
        3 = it is incoherent in many places, but if you spend time reading it, you still can understand the whole story. \\\hline
        4 = it is mostly coherent. Although the text is incoherent in some places, it does not affect reading. \\\hline
        5 = it is completely coherent.\\\hline
    \end{tabular}
    \caption{Rating explanations for coherence.}
    \label{sup:tab:human_evaluation_options_coherence}
\end{table}

\begin{table}[t]
    \centering\small
    \begin{tabular}{|p{0.45\textwidth}|}\hline
        1 = it is completely contradictory to what is described in the table. \\\hline
        2 = it has some facts contradictory to what is described in the table. \\\hline
        3 = it is not supported by the table, and it does not contradict the table. \\\hline
        4 = some of the text is supported by the facts in the table, and the rest of it does not contradict the facts in the table. \\\hline
        5 = it is completely supported by the table.\\\hline
    \end{tabular}
    \caption{Rating explanations for faithfulness.}
    \label{sup:tab:human_evaluation_options_faithful}
\end{table}

\begin{table}[t]
    \centering\small
    \begin{tabular}{|p{0.45\textwidth}|}\hline
        1 = the text is completely irrelevant to the reference. \\\hline
        2 = most of the text is irrelevant to the reference. \\\hline
        3 = some of the text is relevant to the reference. \\\hline
        4 = most of the text is relevant to the reference. \\\hline
        5 = the text is talking about the same thing as the reference.\\\hline
    \end{tabular}
    \caption{Rating explanations for relevance.}
    \label{sup:tab:human_evaluation_options_relevance}
    \vspace{-1em}
\end{table}

\begin{table}[t]
    \centering\small
    \begin{tabular}{|p{0.45\textwidth}|}\hline
        1 = it has quite a few facts contradictory to what is described in the reference. \\\hline
        2 = it has some facts contradictory to what is described in the reference. \\\hline
        3 = it is not supported by the reference, and it does not contradict the reference. \\\hline
        4 = some of the text is supported by the facts in the reference, and the rest of it does not contradict the reference. \\\hline
        5 = it is completely supported by the reference.\\\hline
    \end{tabular}
    \caption{Rating explanations for supportedness.}
    \label{sup:tab:human_evaluation_options_support}
\end{table}

The selected topics for human evaluations are: human (excluding the introduction and biography section), film, single (song), song, album, television series. When evaluating grammaticality and coherence, only the generated text is shown to annotators. The question for grammaticality is ``On a scale of 1-5, how much do you think the text is grammatical? (Note: repetitions are grammatical errors.)'' (option explanations are shown in Table \ref{sup:tab:human_evaluation_options_grammar}), and the question for coherence is ``On a scale of 1-5, how much do you think the text is coherent? (Coherence: Does the text make sense internally, avoid self-contradiction, and use a logical ordering of information?)'' (rating explanations are in Table \ref{sup:tab:human_evaluation_options_coherence}).

When evaluating faithfulness, we show annotators the article data and the generation. The question is ``On a scale of 1-5, how much do you think the text is supported by the facts in the following table?'' (rating explanations are in Table \ref{sup:tab:human_evaluation_options_faithful}).

When evaluating coherence and relevance, annotators were shown the reference text and the generation, as well as the Wikipedia article title and section titles for ease of understanding the texts. Annotators were asked two questions, with one being ``On a scale of 1-5, how much do you think the text is relevant to the reference'' (Table \ref{sup:tab:human_evaluation_options_relevance}), and the other being ``On a scale of 1-5, how much do you think the text is supported by the facts in the reference?'' (Table \ref{sup:tab:human_evaluation_options_support}).

\section{Effect of \entmax}
\label{sec:entmax}

\begin{table}[t]
    \centering
    \small\setlength{\tabcolsep}{3pt}
\begin{tabular}{|l|c|c|c|c|c|}\hline
 & REP & BLEU & PAR-P & PAR-R & PAR-F1 \\\hline
 \multicolumn{6}{c}{Greedy decoding}
\\\hline
base & 38.1 & 14.7 & 61.6 & 27.7 & 35.8 \\
+ ent. + ent. loss & 36.0 & 16.2 & 62.2 & 28.9 & 37.0 \\
+ ent. & 44.5 & 13.9 & 63.5 & 25.5 & 33.9 \\
+ ent. + copy & 43.7 & 14.8 & 64.2 & 26.6 & 35.2 \\
+ copy & 37.8 & 15.8 & 61.3 & 28.3 & 36.3 \\
\hline
 \multicolumn{6}{c}{Beam search (beam size 5)} \\
\hline
base & 33.0 & 15.6 & 66.3 & 28.8 & 37.7 \\
+ ent. + ent. loss & 25.9 & 15.4 & 64.6 & 29.0 & 37.7 \\
+ ent. & 34.7 & 13.8 & 67.2 & 26.6 & 35.8 \\
+ ent. + copy & 34.1 & 15.0 & 69.4 & 28.1 & 37.6 \\
+ copy & 30.1 & 15.9 & 67.1 & 29.4 & 38.5 \\
\hline
\end{tabular}
    \caption{Effect of using \entmax and \entmax loss. When not using the \entmax loss, we use standard cross entropy loss.}
    \vspace{-1em}
    \label{tab:effect_entmax}
\end{table}

In this section, we disentangle the effect of \entmax and that of \entmax loss. We note that (1) when not using the \entmax loss, we use standard cross entropy loss (e.g., in the case of ``base+ent.'' we maximize the log probabilities generated by \entmax); (2) when combining \entmax and copy mechanism, we aggregate the probabilities generated by \entmax and those from softmax. This is because we use the first attention head in the transformer decoder as the copy attention, following the implementation in OpenNMT \cite{klein-etal-2017-opennmt}. While it is feasible to combine the \entmax and \entmax loss with the copy mechanism if we use the sparse transformer \cite{correia-etal-2019-adaptively}, we leave this for future study. We report the results in Table \ref{tab:effect_entmax}. It is interesting to see that when using greedy decoding, ``ent. + ent. loss'' outperforms the baseline model by a significant margin on all the metrics, however the improvement disappears (except for repetition) after we switch to use beam search as the decoding strategy. This is likely because \entmax promotes sparsity in the generated probabilities, making beam search decoding unnecessary. Removing the \entmax loss hurts the performance, but its gains become larger in switching to beam search decoding. Adding copy mechanism improves the performance, leading to comparable performance to the baseline model. Although ``base+ent.+copy'' still underperforms ``base+copy'' when using beam search, we believe that combining \entmax and \entmax loss with the copy mechanism is promising as (1) \entmax is not used in our large models and the initial results have shown that \entmax and the copy mechanism are complementary, so it may further improve our current best performance; (2) \entmax already shows the best performance when using greedy decoding, which has speed and optimization advantages compared to the beam search based decoding strategies especially considering the long-form characteristic of \dataset.

\section{Details of Cyclic Loss}
In this section, we will denote the linearized table where the values are replaced with a special $\langle\text{mask}\rangle$ token by $u_1,\cdots,u_n$, and denote the reference text by $x_1,\cdots,x_m$. Formally, the training loss is
\begin{equation}
    \sum_{w\in S}-\log p(w\vert u_1,\cdots,u_n,\bs{v}_1,\cdots,\bs{v}_m)
\end{equation}

\noindent where $S$ represents the set of masked tokens, and $\bs{v}_1,\cdots,\bs{v}_m$ is the sequence of token-level probabilities predicted by the forward model (in our experiments, these could either come from the softmax function, or the \entmax function).
Specifically, we multiply the backward transformer's input embedding matrix by the $\bs{v}$ probability vectors
to obtain the input representations to the first encoder layer.
We find that it is helpful to add a ``reference loss'' while training with the cyclic loss, %
defined as
\begin{equation}
    \sum_{w\in S}-\log p(w\vert u_1,\cdots,u_n,x_1,\cdots,x_m)
\end{equation}

\noindent This loss does not contain the generation model in it explicitly, but it does lead to an improved backward model by training it with clean inputs. Improving the backward model then increases the benefits of the cyclic loss.\footnote{We experimented with initializing the backward model with pretrained checkpoints, but did not find it helpful.} %

\section{Effect of Article Data and Section Data}
\begin{table}[t]
    \centering
    \small\setlength{\tabcolsep}{5pt}
\begin{tabular}{|l|c|c|c|c|c|}\hline
 & REP & BLEU & PAR-P & PAR-R & PAR-F1 \\\hline
Both & 38.1 & 14.7 & 61.6 & 27.7 & 35.8 \\
Art. only & 60.9 & 8.4 & 55.2 & 14.7 & 20.8 \\
Sec. only & 39.0 & 13.4 & 56.1 & 24.3 & 31.7 \\
\hline
\end{tabular}
    \caption{Effect of dropping section or article data from the input (using the ``base'' setting).}
    \vspace{-0.5em}
    \label{tab:effect_article_section_data_input}
\end{table}

\begin{table}[t]
    \centering
    \small\setlength{\tabcolsep}{5pt}
\begin{tabular}{|l|c|c|c|c|c|}\hline
 & REP & BLEU & PAR-P & PAR-R & PAR-F1 \\\hline
None & 37.8 & 15.8 & 61.3 & 28.3 & 36.3 \\
Both & 35.9 & 15.8 & 62.0 & 28.5 & 36.7 \\
Art. only & 37.2 & 15.8 & 61.7 & 28.1 & 36.2 \\
Sec. only & 34.8 & 15.9 & 61.9 & 28.2 & 36.2 \\\hline
\end{tabular}
    \caption{Effect of dropping section or article data when using cyclic training. The results are based on the ``base + copy'' and ``base + copy + cyclic loss'' settings.}
    \vspace{-0.5em}
    \label{tab:effect_article_section_data_mask}
\end{table}

We report results in Table~\ref{tab:effect_article_section_data_input} for the models that are trained with partial data input, where art.~only and sec.~only indicate that we use only article data or section data, respectively. We always use title data.
Section data contributes the most to the BLEU and PAR scores, but using section data and article data together is the best setting.

We also investigate the effect of partial data input for the cyclic loss in Table~\ref{tab:effect_article_section_data_mask}, where ``None'' is the model that is not trained with the cyclic loss. We note that in this setting, we still use both data resources as the input to the forward model, but vary the input data and the gold standard for the backward model. Although using only section data gives the best REP score and improves the PAR-P score, it does not help the model in other metrics.
Combining the article data with the section data gives significant improvements to the PAR-F1 score compared to section data alone.

Both experiments show that there are interactions between these two data resources that can help models to learn better from both kinds.

\begin{figure*}[t]
    \centering
    \includegraphics[scale=0.4]{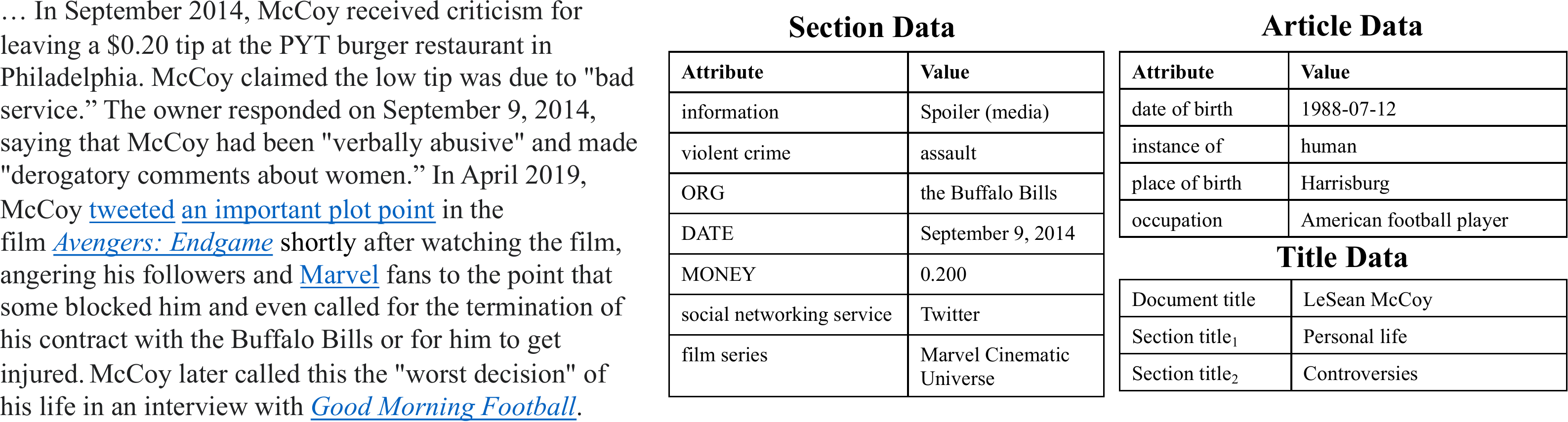}
    \caption{An example from \dataset. Only parts of the tables and reference are shown due to space constraints. Underlined texts are hyperlinks. Records with the attributes ``DATE'', ``ORG', ``MONEY'' are from NER. Subscripts for section titles indicate the ordering of nesting, where smaller numbers indicate higher level sections.}
    \label{fig:table}
\end{figure*}

\begin{table*}[t]
\centering
\tiny
\setlength{\tabcolsep}{4pt}
\begin{tabular}{|p{0.05\textwidth}|p{0.91\textwidth}|}\hline
\multicolumn{1}{|l|}{Method} &\multicolumn{1}{c|}{Generated Text} \\\hline

sampling 0.9 & In a two-issue mini-series titled "Messiah Complex", portrayed by Maisie Williams, Father Craig is ordered to rescue the world from Reavers and run out of the X-Force causing the X-Force to permanently withdraw. The original humans were replaced by slots and Muir Island being pulled away. They also possessed the entire X-Force after issue \#9 in which the X-Force appeared. Claremont and Sinclair later reprised their parts in "The Donderful World of Wolverine" with Wolfsbane as one of the new X-Men and her group taken back to their reservation for unknown reasons. \\\hline
beam search (size=5) & In the "Messiah Complex" storyline, Wolfsbane is a member of the X-Men. She is a member of the X-Factor Investigations team and a member of the X-Force/X-Force/X-Force/X-Force/X-Force/X-Force/X-Force/X-Force/X-Force/X-Force/X-Force/X-Force/X-Force/X-Force/X-Force/X-Force/X-Force/X-Force/X-Force/X-Force/X-Force/X-Force/X-Force/X-Force/X-Force/X-Force/X-Force/X-Force/X-Force/X-Force/X-Force/X-Force/X-Force/X-Force/X-Force/X-Force/X-Force/X-Force/X-Force/X-Force/X-Force/X-Force/X-Force/X-Force/X-Force/X-Force/X-Force/X-Force/X-Force/X-Force/X-Force/X-Force/X-Force/X-Force/X-Force/X-Force/X-Force/X-Force/X-Force/X-Force/X-Force/X-Force/X-Force/X-Force/X-Force/
\\\hline
blocking (size=5) & In the "Messiah Complex" storyline, Wolfsbane is a member of the X-Men, a group of mutants recruited by Father Craig to fight the Reavers in Muir Island. Father Craig is killed in the battle with Reavers, but Father Craig manages to rescue the mutant from the Rain of Gesham and retrieve the mutant. She is later revealed to have been killed by Rain in a battle with Father Craig. In the aftermath, Father Craig tells Father Craig that she is the only mutant to be killed by a mutant. \\\hline
\hline

sampling 0.9 & Journey to the Center of the Earth is a 1959 American science fiction film based on Jules Verne's 1959 novel of the same name. The film, directed by Henry Levin, stars James Mason as Professor Otto Lidenbrock, a scientist who gets entangled in the power of the archenemy as he tries to impress his closest allies and compromises them with the future. The film marks the first screen appearance of future CinemaScope of 20th Century Fox lead singer Peter Ronson. "The 20th Century Fox" earned critical acclaim and was a box-office success, grossing \$10 million worldwide against a production budget of \$3.44 million . \\\hline
beam search (size=5) & Journey to the Center of the Earth is a 1959 American science fiction film directed by Henry Levin and starring James Mason, Pat Boone, and Arlene Dahl. The screenplay by Charles Brackett and Walter Reisch is based on Jules Verne's novel "A Journey to the Center of the Earth" (1959). The film was produced by Charles Brackett and Walter Reisch and distributed by Twentieth Century-Fox Film Corporation. The film was nominated for an Academy Award for Best Art Direction. It was also nominated for an Academy Award for Best Art Direction and a Golden Globe Award for Best Visual Effects. It was also nominated for a Golden Globe Award for Best Motion Picture - Musical or Comedy, but lost to 20th Century Fox's "CinemaScope". \\\hline
blocking (size=5) & Journey to the Center of the Earth is a 1959 American science fiction film directed by Henry Levin and starring James Mason, Pat Boone, and Arlene Dahl. The screenplay by Charles Brackett and Walter Reisch is based on Jules Verne's novel of the same name. The film was produced by CinemaScope and 20th Century Fox and distributed by Twentieth Century-Fox Film Corporation and Joseph M. Schenck Enterprises, Inc. It was nominated for an Academy Award for Best Art Direction. \\\hline
\hline

sampling 0.9 & In May 2013, McCoy allegedly assault a Spoiler business official. The incident motivated him to take a business trip to Harrisburg, Pennsylvania and sign a letter from the Buffalo Bills paying him over \$50,000 as well as verbally prohibiting legal kickoff work. In September 2014, McCoy shared the issue with some former players of the first two film franchises. McCoy played a leading role in the 2012 Marvel Cinematic Universe film "Erab of the Press", and was named one of the film's biggest stars on Twitter. On September 9, 2014, McCoy filed a lawsuit against PYT to change its Twitter name to LeSean McCoy Productions. \\\hline
beam search (size=5) & On September 9, 2014, McCoy was arrested in Harrisburg, Pennsylvania on suspicion of assault. He was charged with assault and battery. In May 2013, McCoy was fined over \$50,000 by the Buffalo Bills. In September 2014, McCoy was arrested in Harrisburg, Pennsylvania on suspicion of assaulting a woman who had been raped by McCoy. McCoy was charged with assault and possession of marijuana. McCoy was suspended from the PYT for the first two games of the Marvel Cinematic Universe. \\\hline
blocking (size=5) & On September 9, 2014, McCoy was arrested in Harrisburg, Pennsylvania on suspicion of assault. He was charged with assault and battery. In May 2013, he was fined over \$50,000 by the Buffalo Bills. In September 2014, he was suspended for two games by the PYT for violating the Marvel Cinematic Universe. He was released by the Bills in October of the same year. He was cleared of all charges on Twitter, and was banned from playing in the 2014 Pro Bowl due to his Twitter account. \\\hline
\end{tabular}

    \caption{Top: generation examples for the first instance in Figure~\ref{fig:dataset-example}. Middle: generation examples for the second instance in Figure~\ref{fig:dataset-example}. Bottom: generation examples that correspond to the instance in Figure \ref{fig:table}.}
    \label{tab:generation_examples_appendix}
\end{table*}

\section{Generation Examples}

We show the full set of generations in Table \ref{tab:generation_examples_appendix}. The part of input data and reference text for Table \ref{tab:generation_examples_appendix} is shown in Figure \ref{fig:table}.

\section{Details of Decoding Strategies}

\paragraph{Nucleus Sampling.} Generating long sequences usually suffers from repetitions. Nucleus sampling \cite{Holtzman2020The} aims to reduce the repetitions in generations by sampling from truncated probability distributions. The truncation is based on whether the cumulative probability is above a threshold. We set the threshold to be 0.9 as suggested in \citet{Holtzman2020The}.

\paragraph{Beam Search with $n$-gram Blocking.} \citet{paulus2018a} found it effective to reduce the repetitions during beam search by ``blocking'' $n$-grams that have been generated in previous decoding steps. We follow their approach by using trigram blocking and setting the probability of repeated trigrams to be 0 during beam search.